\documentclass[runningheads]{llncs}
\usepackage[T1]{fontenc}
\usepackage{graphicx,verbatim}
\usepackage{amsmath}
\usepackage{amssymb}
\usepackage{xspace}
\usepackage[table,xcdraw,dvipsnames]{xcolor}
\usepackage{hyperref}

\usepackage{booktabs}
\usepackage{makecell}

\newcommand{\figref}[1]{Fig.~\ref{#1}}
\newcommand{\tabref}[1]{Table~\ref{#1}}
\newcommand{\ie}{\emph{i.e.}}
\newcommand{\eg}{\emph{e.g.}}
\newcommand{\wrt}{w.r.t.}

\begin{document}
\title{Aligning Fetal Anatomy with Kinematic Tree Log-Euclidean PolyRigid Transforms}
\titlerunning{Aligning Fetal Anatomy}
\author{Yingcheng Liu\inst{1} \and
Athena Taymourtash\inst{1} \and
Yang Liu\inst{2} \and
Esra Abaci Turk\inst{3} \and
William~M.~Wells\inst{1} \and
Leo Joskowicz\inst{4} \and
P. Ellen Grant\inst{3} \and
Polina Golland\inst{1}}
\authorrunning{Y. Liu et al.}
%
\institute{Computer Science and Artificial Intelligence Lab, MIT, Cambridge, USA \and
California Institute of Technology, Pasadena, USA \and
Boston Children’s Hospital and Harvard Medical School, Boston, USA \and
The Hebrew University of Jerusalem, Jerusalem, Israel\\
\email{liuyc@mit.edu}}

\maketitle              

\begin{abstract}

Automated analysis of articulated bodies is crucial in medical imaging.
Existing surface-based models often ignore internal volumetric structures and rely on deformation methods that lack anatomical consistency guarantees.
To address this problem, we introduce a differentiable volumetric body model based on the Skinned Multi-Person Linear (SMPL) formulation, driven by a new Kinematic Tree-based Log-Euclidean PolyRigid (KTPolyRigid) transform.
KTPolyRigid resolves Lie algebra ambiguities associated with large, non-local articulated motions, and encourages smooth, bijective volumetric mappings.
Evaluated on 53 fetal MRI volumes, KTPolyRigid yields deformation fields with significantly fewer folding artifacts.
Furthermore, our framework enables robust groupwise image registration and a label-efficient, template-based segmentation of fetal organs.
It provides a robust foundation for standardized volumetric analysis of articulated bodies in medical imaging.


\end{abstract}

\section{Introduction}
\label{sec:intro}

Automated shape and motion analysis of articulated bodies is critical in medical imaging.
In prenatal care, fetal biometric assessment serves as the established clinical gold standard for evaluating normative growth trajectories~\cite{papageorghiou2014international}.
Quantification of fetal motion in MRI is emerging as a potential biomarker of fetal health and neuromuscular development~\cite{vasung2023cross}.

A common approach to automating this analysis involves building parametric body models and aligning them to image 
data~\cite{hesse2019learning,liu2025fetuses}.
Existing approaches often employ surface models primarily developed for real-time character animation, where visual plausibility is prioritized.
In biomedical image analysis, volumetric models coupled with differentiable, smooth, and bijective deformations are preferred.
Volumetric modeling enables us to analyze rich internal anatomical information.
Smooth bijective mappings produce faithful image transformations without tearing or folding.
A differentiable mapping supports gradient computation \wrt~model parameters and facilitates image registration.

Extending surface models to volumetric or biomechanical representations presents several challenges.
Tetrahedral mesh-based volumetric models require careful handling to avoid mesh inversion and are difficult to tailor to be end-to-end differentiable~\cite{abulnaga2019placental,kim2017data}.
Biomechanically accurate models produce more realistic body motion but often omit essential soft tissues and organs~\cite{keller2023skin}.
The Log-Euclidean PolyRigid (PolyRigid) transform offers a principled method for modeling volumetric data with articulated structure but assumes small local deformations~\cite{arsigny2009fast}.
Large, non-local motions can lead to ambiguity in mapping between the Lie group and Lie algebra, limiting its application in fetal body modeling.

In this paper, we propose a novel Kinematic Tree Log-Euclidean PolyRigid (KTPolyRigid) transform, alongside a differentiable volumetric body model based on the Skinned Multi-Person Linear (SMPL) formulation~\cite{loper2023smpl}.
KTPolyRigid generalizes PolyRigid to handle large, non-local motions by using the body's kinematic tree. 
The key insight is that although global body part transformations may be large, relative transformations between adjacent parts in the kinematic tree typically remain small.
By averaging relative rotations rather than global ones, we avoid Lie algebra mapping ambiguities. 

Our contributions are threefold.
First, we introduce KTPolyRigid, a novel method for smooth, bijective volumetric deformation of articulated bodies.
It yields deformation fields with improved regularity and fewer folding artifacts compared to baseline methods.
Second, we propose a groupwise image registration method that leverages our differentiable volumetric model to improve the anatomical alignment across subjects.
We employ our method to create the first population average of fetal anatomy in a canonical pose.
Third, we illustrate the utility of our framework by demonstrating label-efficient template-based segmentation of fetal lungs.

\begin{figure}[t]
\centering
\includegraphics[width=\linewidth]{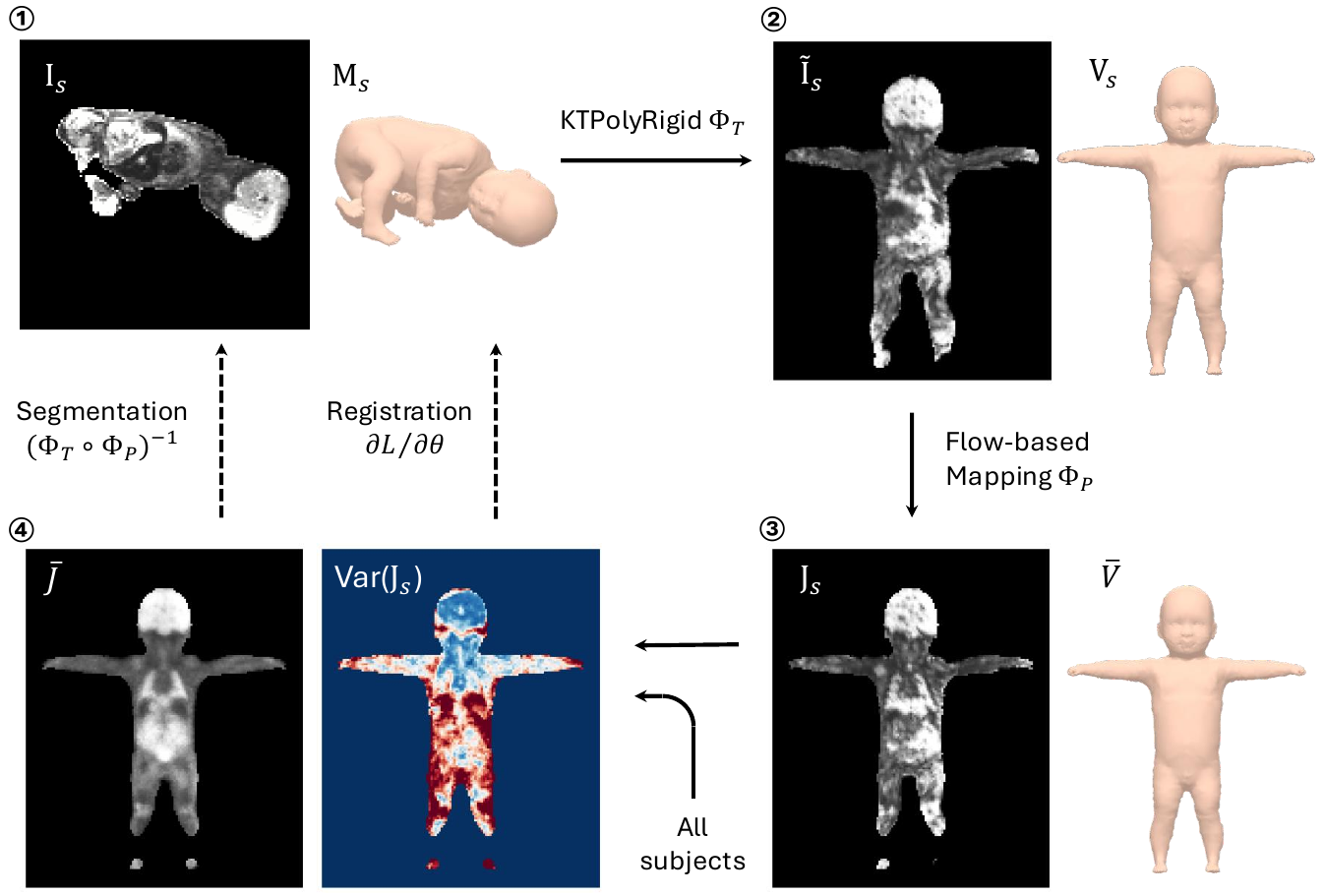}
\caption{
    \textbf{Method overview.}
    KTPolyRigid $\Phi_T$ flattens the fetal body into a canonical T-pose. 
    The flow-based mapping $\Phi_P$ further standardizes the body shape to the population mean $\bar{V}$.
    Averaging the population-level canonical images $J_s$ produces a population mean image $\bar{J}$.
    Groupwise registration optimizes the body pose $\theta$ to minimize image variance across subjects.
    Segmentation is achieved by total transformation $\Phi = \Phi_T \circ \Phi_P$.
}
\label{fig:method-2}
\end{figure}

\section{Method}
\label{sec:method}

\subsubsection{SMPL Model.}

We build our volumetric body model based on a surface-based articulated statistical shape model SMPL~\cite{loper2023smpl}.
It consists of a fixed-topology surface mesh with $N=6890$ vertices and $F=13796$ faces, controlled by a kinematic tree with $K=23$ joints.
The body pose is parameterized by $\theta \in \mathbb{R}^{3 \times K}$ with each joint rotation represented as an angle-axis vector.
SMPL uses T-pose as the canonical pose~$\theta^*$ to standardize subject shapes.
PCA is applied to a dataset of T-pose shapes, yielding a population mean shape $\bar{V} \in \mathbb{R}^{3N}$ and principal components (PCs) $B_s(\beta): \mathbb{R}^{|\beta|} \to \mathbb{R}^{3N}$, where $|\beta|=10$ is the number of PCs used to model the body shape.
A new subject's canonical shape is specified as $V_s(\beta) = \bar{V} + B_s(\beta)$ using a subject-specific shape vector $\beta$.
To deform $V_s$ into an arbitrary body mesh $M_s$ with a target pose $\theta$, SMPL employs Linear Blend Skinning (LBS).
Specifically, the pose-blend-shape $B_p(\theta): \mathbb{R}^{9 \times K} \to \mathbb{R}^{3N}$ is a linear function of the body pose~$\theta$.
$B_p(\theta)$ offsets the subject's canonical shape $V_s$ before LBS to account for pose-dependent soft tissue deformations.
LBS weighting functions $\mathcal{W} \in \mathbb{R}^{N \times K}$ are defined on the surface vertices. 
Element $w_{i,k} \geq 0$ of $\mathcal{W}$ indicates the influence of joint $k$ on vertex $i$, subject to the partition of unity constraint $\sum_{k=1}^K w_{i,k} = 1$ for all $i$.

\subsubsection{From Images to a Canonical Shape Model.}

Let $I_s : \mathbb{R}^3 \to \mathbb{R}$ denote the original image (\eg,~a volumetric MRI scan) of subject $s$.
The surface mesh~$M_s$ is fit to the image $I_s$ using landmarks and body segmentations~\cite{liu2025fetuses}.
We assume initial estimates of pose $\theta$ and shape $\beta$ are obtained from this alignment step.
While the SMPL model defines a surface, we develop volumetric modeling to map the interior tissue. 
Let $\Omega \subset \mathbb{R}^3$ denote the continuous interior volumetric domain enclosed by the subject's surface $\partial \Omega$, discretized by the mesh $M_s$.
A volumetric map $\Phi_T(\cdot; \beta, \theta): \mathbb{R}^3 \to \mathbb{R}^3$ warps the native image $I_s$ into a subject-specific canonical image $\tilde{I}_s = I_s \circ \Phi_T$ that visualizes the subject's body in a standardized T-pose, which enables intrasubject comparison of anatomical structures.
To model intrinsic body shape variation, we define transformation $\Phi_P(\cdot; \beta): \mathbb{R}^3 \to \mathbb{R}^3$ that maps the subject-specific canonical shape $V_s$ to the population-level mean shape $\bar{V}$.
This step yields the population-level canonical image $J_s = \tilde{I}_s \circ \Phi_P$ for each subject.
All population-level images $J_s$ share the same shape $\bar{V}$ and pose $\theta^*$.
Population-level mean image $\bar{J}$ is simply the voxelwise average of all population canonical images $J_s$ that facilitates population-level modeling of image data.

In this paper, we define volumetric mappings $\Phi_T$ and $\Phi_P$ that are smooth, bijective, and differentiable w.r.t.~the body pose $\theta$.
In the remainder of this section, we develop the construction of $\Phi_T$ and $\Phi_P$, and demonstrate how they enable groupwise image registration and template-based anatomical segmentation.

\subsubsection{Kinematic Tree Log-Euclidean PolyRigid Transform $\Phi_T$.}

SMPL model parameterizes the body deformation using a group of rigid transformations~$\{\mathbf{T}_k\}$ with each $\mathbf{T}_k \in SE(3)$ defining a rigid part in the model.
The Log-Euclidean PolyRigid framework fuses such rigid transformations into a smooth volumetric mapping~\cite{arsigny2009fast,martin2009log,legouhy2023polaffini}.
The general framework builds a stationary velocity field by averaging the principal logarithms of the rigid parts in $\mathfrak{se}(3)$ and obtains the transformation as the flow of this field~\cite{arsigny2009fast}.
This guarantees diffeomorphism through ODE integration.
For computational efficiency, a closed-form approximation directly takes the exponential of the spatially weighted log-transformations~\cite{gopalakrishnan2025polypose}.
We adopt this closed-form approximation because our application involves a large number of volumes and gradient-based groupwise registration, as we will describe in the later sections.
Using the general ODE-based formulation would incur prohibitive computational cost.
Formally, the closed-form transformation $\mathbf{T}(x)$ of a 3D point $x$ is given by
\begin{equation}
    \mathbf{T}(x) = \exp \left\{
        \sum_{k=1}^{K} w_k(x) \log(\mathbf{T}_k) \right\},
\label{eq:polyrigid}
\end{equation}
where $w_k: \mathbb{R}^3 \to [0,1]$ are continuous, spatially varying weighting functions with $\sum_{k=1}^K w_k(x) = 1$ for all $x$.
Operators $\log(\cdot)$ and $\exp(\cdot)$ denote the Lie group logarithm and exponential maps respectively.
However, this formulation assumes the deformation is small and local to the identity.
The principal logarithm is only well-defined within the injectivity radius of $SE(3)$.
The mapping becomes ambiguous when the rigid transformations are large, \ie,~multiple Lie algebra elements map to the same Lie group element.
This precludes direct application of PolyRigid to body models, where body parts can undergo large, non-local motion.

Is there a way to disambiguate the mapping?
We observe that while body parts can undergo large transformations, the relative motion between connected parts is small and is always constrained by the underlying kinematic tree.
In constrast to PolyRigid that uses the tangent space at the identity to perform averaging, we define a local tangent space for each voxel using a spatially varying reference rigid transformation $\tilde{\mathbf{T}}(x)$.
The relative transformations $\Delta \mathbf{T}_k(x) = \tilde{\mathbf{T}}(x)^{-1} \mathbf{T}_k$ are often small and within the injectivity radius. 
This pulls the transformations back into a local neighborhood where the logarithm is unique and computationally stable.
More formally, the KTPolyRigid transform $\Phi_T$ is defined as:
\begin{equation}
\Phi_T (x) = \tilde{\mathbf{T}}(x) \circ
    \exp \left\{\sum_{k=1}^{K} w_k(x) \log \left( \tilde{\mathbf{T}}(x)^{-1} \mathbf{T}_k \right) \right\}.
\end{equation}
In practice, we select the reference rigid transformation $\tilde{\mathbf{T}}(x)$ as the one with the largest blending weight at voxel $x$: $\tilde{\mathbf{T}}(x) = \mathbf{T}_{\tilde{k}(x)}$ where $\tilde{k}(x) = \arg\max_k w_k(x)$.
Although argmax itself is discontinuous, $\Phi_T$ remains continuous across the boundary between adjacent regions.
As $x$ approaches the boundary, two weights both approach $1/2$, so $\Phi_T$ smoothly converges to the tangent-space average of the two rigid transforms regardless of which one is the reference $\tilde{\mathbf{T}}(x)$.
The mapping is non-differentiable only on a measure-zero set similar to the derivatives of distance transforms.

SMPL defines weighting functions $w_k$ only on the surface vertices. 
To obtain a volumetric function, we extend these weights into the interior $\Omega$ by minimizing the Laplacian energy $E(\mathbf{w}) = \int_{\Omega} \sum_{k} \| \nabla w_k(x) \|^2 dx$, subject to partition of unity, non-negativity, and Dirichlet boundary constraints~\cite{kim2017data}.
We discretize this on a voxel grid and solve using projected gradient descent.

\subsubsection{Flow-Based Diffeomorphic Mapping $\Phi_P$.}

The KTPolyRigid transform $\Phi_T$ applies to an input image $I_s$ and produces a canonical image $\tilde{I}_s$ with the fetal body in the standard T-pose.
To obtain population-level canonical image $J_s$, we construct a flow-based diffeomorphic mapping based on our statistical shape model.
Let $\Omega(0) \subset \mathbb{R}^3$ denote the initial interior domain.
We define a linear path in the shape parameter space from the subject $\beta_s$ to the population mean $\beta_{\text{pop}} = \mathbf{0}$, given by $\beta(t) = (1-t)\beta_s + t\beta_{\text{pop}}$ for $t \in [0, 1]$.
This induces a continuous deformation of the boundary surface, where the vertices of $\partial \Omega(t)$ are given by $\mathbf{V}(t) = \bar{V} + B_S(\beta(t))$. 
Because $B_s$ is linear \wrt~the shape parameters, the velocity of the boundary vertices is constant: $\dot{\mathbf{V}}(t) = B_S(\beta_{\text{pop}} - \beta_s)$.
We extend this boundary velocity to the interior volume $\Omega(t)$ using generalized barycentric coordinates.
Specifically, we use Mean Value Coordinates (MVC) to define a volumetric velocity field $\mathbf{v}: \Omega \times [0,1] \to \mathbb{R}^3$~\cite{ju2023mean}:
$$\mathbf{v}(x, t) = \sum_{n=1}^{N} \psi_n(x; \partial \Omega(t)) \cdot \dot{\mathbf{V}}_n(t),$$
where $\psi_n$ represents the MVC weight associated with the vertex $n$ of the evolving boundary $\partial \Omega(t)$ and $\dot{\mathbf{V}}_n(t)$ is the velocity of that specific vertex.
The smooth, bijective volumetric mapping $\Phi_P: \Omega(0) \to \Omega(1)$ is defined as the flow of this velocity field, characterized by the ODE $\frac{d}{dt} \Phi_P(x, t) = \mathbf{v}(\Phi_P(x, t), t)$ with the initial condition $\Phi_P(x, 0) = x$.
We solve the ODE numerically using explicit Euler integration.

\subsubsection{Groupwise Registration and Template-Based Segmentation.}

Composing the KTPolyRigid transform $\Phi_T$ and the flow-based mapping $\Phi_P$, we obtain dense correspondences that map the population canonical space to the subject's native space: $\Phi(x) = \Phi_T \circ \Phi_P$.
We transfer template segmentations to the subject space using the inverse transformation $\Phi^{-1}$.
To create $J_s$, we sample the native scan images $I_s$ onto a regular grid in the population-level canonical space using $\Phi$ and trilinear interpolation.

We can further refine the alignment of internal anatomical structures through groupwise image registration~\cite{zollei2005efficient}.
This amounts to optimizing the body pose parameters $\theta$ to minimize variance across subjects.
We introduce a local perturbation in the tangent space $\mathfrak{se}(3)$ around the initial rigid transformations.
Let $\mathbf{T}_{s,k} \in SE(3)$ denote the initial rigid transformation for joint $k$ of subject $s$.
We parameterize the residual motion using a twist vector $\xi_{s,k} \in \mathbb{R}^6$, comprising rotational $\omega$ and translational $v$ components.
We then linearize the exponential map to improve computational efficiency, defining a refined rigid transformation with the matrix representation:
\begin{equation}
\mathbf{T}_{s,k}^{\prime}(\mathbf{T}_{s,k}, \xi_{s,k})
    = \exp(\xi_{s,k}) \cdot \mathbf{T}_{s,k}
    \approx \left( \mathbf{I} + \widehat{\xi}_{s,k} \right) \cdot \mathbf{T}_{s,k}
    = \left( \mathbf{I} + \begin{bmatrix} [\omega]_{\times} & v \\ 0 & 0 \end{bmatrix} \right) \cdot \mathbf{T}_{s,k},
\end{equation}
where $\widehat{\xi} \in \mathfrak{se}(3)$ is the matrix representation of the twist, and $[\omega]_{\times}$ is the skew-symmetric matrix of $\omega$.
We optimize these twist parameters to minimize the intensity variance across the population over the interior volume domain $\Omega$, while regularizing the magnitude of the perturbations to be small.
Formally, we minimize an objective function
\begin{equation}
\mathcal{L}(\{\xi_{s,k}\})
    = \int_{\Omega} \sum_{s=1}^{S} \left\| I_s(\Phi(x)) - \bar{J}(x) \right\|^2 dx + \lambda \sum_{s,k} \|\xi_{s,k}\|^2,
\end{equation}
where $\bar{J}(x) = \frac{1}{S} \sum_{s=1}^{S} I_s(\Phi(x; \{\mathbf{T}_{s,k}^{\prime}(\mathbf{T}_{s,k}, \xi_{s,k})\}))$ is the population mean image at spatial location $x$.
This step minimizes the spatial intensity variance and produces sharper population mean image $\bar{J}$.

\section{Experimental Results}
\label{sec:experiment}

\begin{figure}[t]
\centering
\includegraphics[width=\linewidth]{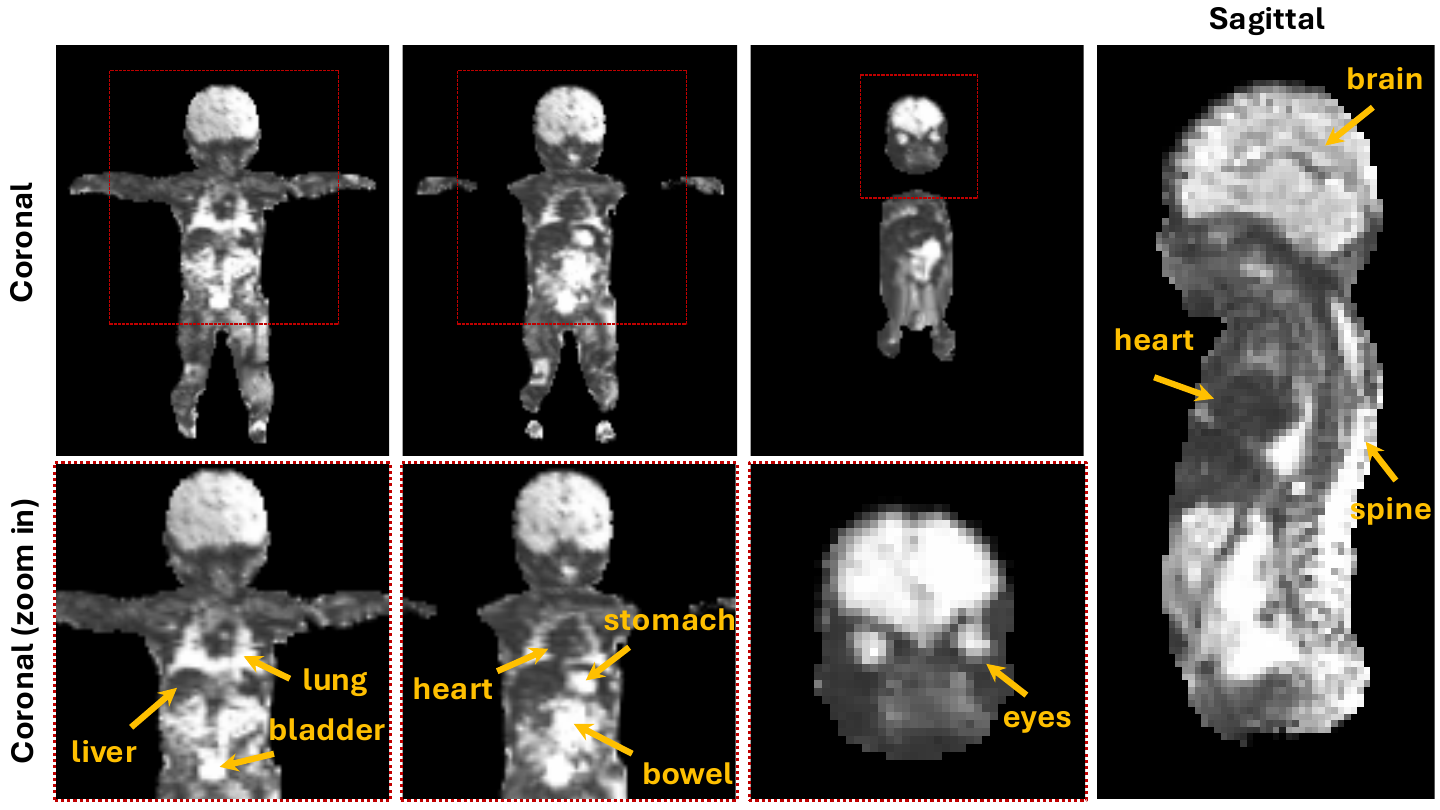}
\caption{
\textbf{Example canonical image $\tilde{I}_s$ of a subject $s$.}
Left three panels: Three coronal cross-sections (full and zoomed-in views).
Right panel: mid-sagittal cross-section.
}
\label{fig:exp-1}
\end{figure}

\subsubsection{Data.}
We evaluate our method on 53 healthy singleton 3D fetal MRI volumes (3T Siemens Skyra, EPI, $3 \times 3 \times 6 \text{mm}^3$ voxel size).
For initial registration, the surface body model is aligned to anatomical landmarks and segmentation masks~\cite{liu2025fetuses,xu2021motion}.

\subsubsection{Baseline Methods.}
We compare KTPolyRigid against two baseline methods: 
(i)~volumetric LBS deformation field is a convex combination of $\mathbf{T}_k$ with matrix representation $\Phi_{\text{LBS}} = \sum_{k=1}^K w_k(x) \mathbf{T}_k$,
and (ii) the closed-form PolyRigid approximation defined in \eqref{eq:polyrigid}.
All method share the weighting functions $w_k$.

\subsubsection{Canonical Image $\tilde{I}_s$ Visualization.}
\figref{fig:exp-1} shows an example image $\tilde{I}_s$ in the canonical T-pose.
We observe anatomical structures, such as the brain, eyes, lung, heart, and liver, in the coronal view.
On the sagittal view, the entire spine appears in the same cross-section, facilitating the visualization and analysis.

\subsubsection{Deformation Field $\Phi_T$ Evaluation.}

\begin{figure}[t]
\centering
\includegraphics[width=\linewidth]{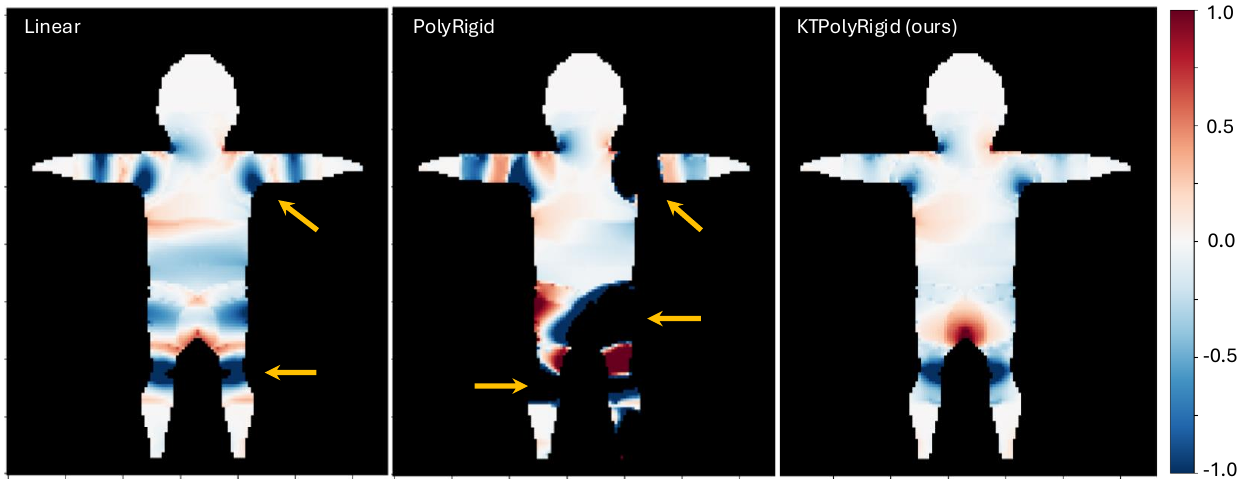}
\caption{
    \textbf{Local distortions.}
    Spatial log determinant Jacobian field $\log_2 \det \left[\partial_x \Phi_T(x) \right]$
    of the deformation $\Phi_T$ for subject in \figref{fig:exp-1}.
}
\label{fig:exp-jacobian}
\end{figure}

\begin{table}[t]
\footnotesize
\centering
\caption{\textbf{Deformation field statistics.} Mean $\pm$ standard deviation are reported. ($\downarrow$) indicates that lower values represent better performance.}
\label{tab:deformation_results}
\renewcommand{\arraystretch}{1.2}
\setlength{\tabcolsep}{3.5pt}
\begin{tabular}{lcccc}
\toprule
\textbf{Method} & \textbf{\% folds} ($\downarrow$) & \makecell{$\mathbf{STD \log_2 |\det J|}$ \\ ($\downarrow$)} & \makecell{\textbf{GPU Time} \\ (ms, $\downarrow$)} & \makecell{\textbf{GPU Mem.} \\ (MB, $\downarrow$)} \\
\midrule
LBS          & $1.88 \pm 0.57$ & $0.89 \pm 0.08$ & $0.33 \pm 0.03$ & $24.27 \pm 0.95$ \\
PolyRigid~\cite{arsigny2009fast} & $5.37 \pm 3.85$ & $0.90 \pm 0.18$ & $93.50 \pm 6.13$    & $3332 \pm 129$   \\
KTPolyRigid (ours) & $1.38 \pm 0.40$ & $0.74 \pm 0.06$ & $31.19 \pm 2.69$    & $1551 \pm 66$    \\
\bottomrule
\end{tabular}
\end{table}

To assess the regularity of the mappings, we compute the determinant of the Jacobian $\det \left[\partial_x \Phi_T(x) \right]$ (\figref{fig:exp-jacobian}).
Regions with $\det \left[\partial_x \Phi_T(x) \right] \leq 0$ indicate local folding and loss of bijectivity, while values close to $1$ represent smooth, physically plausible deformations.
LBS produces artifacts near the shoulder and hip joints.
PolyRigid produces severe artifacts due to incorrect Lie algebra mapping when large transformations are involved.
In contrast, KTPolyRigid produces a smooth deformation field with fewer folding artifacts.
\tabref{tab:deformation_results} reports quantitative metrics for the entire cohort.
KTPolyRigid yields reduced folding artifacts and improved smoothness.

\subsubsection{Groupwise Image Registration.}

\begin{figure}[t]
\centering
\includegraphics[width=\linewidth]{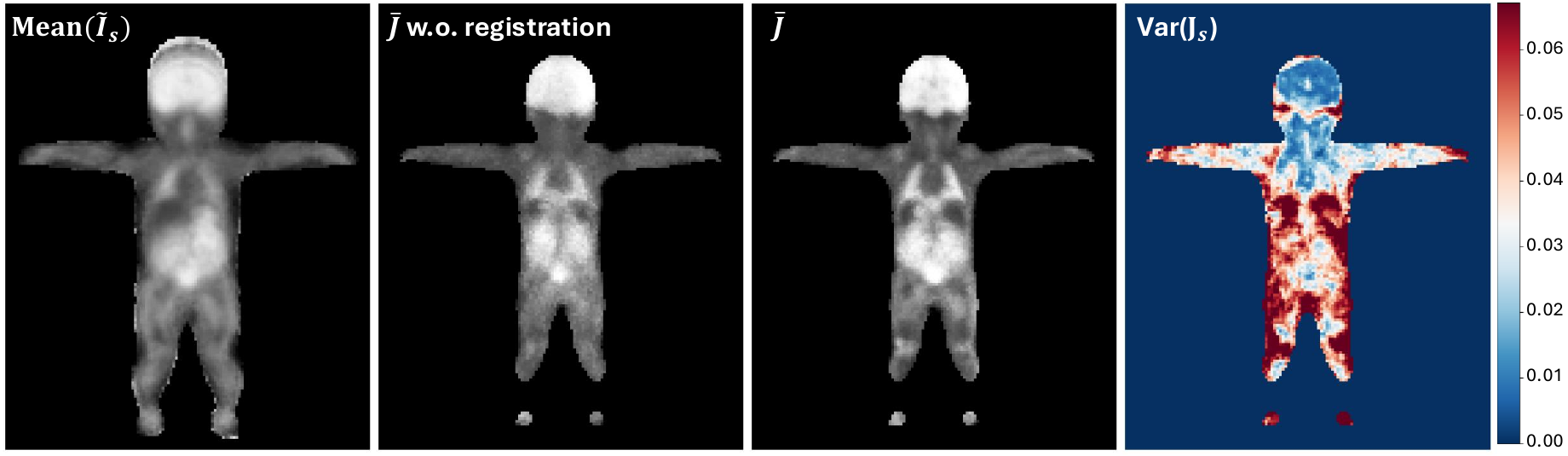}
\caption{
    \textbf{Population-level mean image $\bar{J}$.}
}
\label{fig:exp-2}
\end{figure}

\figref{fig:exp-2} visualizes the population-level mean image $\bar{J}$ before and after groupwise registration.
Simply averaging $\tilde{I}_s$ across subjects produces a blurry image with artifacts near boundaries due to variation in the body shape.
Using $\Phi_P$ standardizes the body shape and reduces boundary artifacts.
Groupwise image registration further improves the alignment of anatomy, producing a sharper population mean image $\bar{J}$.

\subsubsection{Fetal Lung Segmentation.}

To evaluate our framework's utility for segmentation, we trained a U-Net~\cite{ronneberger2015u} on the population canonical images $J_s$ to predict fetal lung segmentation.
In a 5-fold cross-validation experiment, we obtained an average Dice score of $0.81 \pm 0.02$ without hyperparameter tuning.
Notably, if we reduce the training set to be a random subset of $5$ subjects, the model still achieves a competitive Dice score of $0.76 \pm 0.06$.
This is largely due to the high spatial consistency of lung positioning across subjects in the canonical space $J_s$.
Even a model predicting a population-average lung mask yields reasonable Dice score.
This demonstrates that mapping scans to a population canonical space simplifies the localization of anatomical structures and significantly improves efficiency for anatomical segmentation.

\section{Conclusion}

In this paper, we presented KTPolyRigid, a Kinematic Tree-based Log-Euclidean PolyRigid transform for differentiable volumetric articulated body modeling.
Our method resolves the ambiguity of Lie group mappings for large, non-local deformations, encouraging smooth and bijective volumetric mappings.
We demonstrated our approach on fetal MRI data, showing improved deformation regularity and enabling robust groupwise registration and label-efficient segmentation.
Standardizing fetal anatomy in a population-level canonical space facilitates population analysis and provides a foundation for more detailed clinical assessments.

\subsubsection{Disclosure of Interests.}
The authors have no competing interests to declare.

\subsubsection{Acknowledgements.}
This work has been funded by NIH NIBIB 1R01EB036945, NIH NICHD 1R01HD114338, NIH NIBIB 1R01EB032708 and MIT-CSAIL Wistron Program.
We are grateful to the members of the MIT Medical Vision Group and the members of FNNDSC at Boston Children's Hospital for helpful discussions.

\bibliographystyle{splncs04}
\bibliography{bibliography}

\end{document}